\documentclass[11pt,a4paper]{article}
\usepackage{authblk}
\usepackage{blindtext}
\usepackage[hyperref]{ranlp2023}
\usepackage{times}
\usepackage{amsmath}
\usepackage{algorithm}
\usepackage{algpseudocode}
\usepackage{float}

\usepackage{graphicx}
\usepackage{tabularx,booktabs}

\usepackage{latexsym}

\DeclareMathOperator*{\argmax}{arg\,max}

\usepackage{microtype}

\aclfinalcopy % Uncomment this line for the final submission
\title{TreeSwap: Data Augmentation for Machine Translation via Dependency Subtree Swapping}

\author[1]{Attila Nagy}
\author[1,2]{Dorina Lakatos}
\author[1,2]{Botond Barta}
\author[2]{Judit Ács}
\affil[1]{Department of Automation and Applied Informatics \protect\\
Budapest University of Technology and Economics}
\affil[ ]{\texttt{attila.nagy234@gmail.com}}
\affil[2]{Institute for Computer Science and Control}
\affil[ ]{\texttt{\{botondbarta, dorinalakatos, acsjudit\}@sztaki.hu}}

\date{}

\begin{document}
\maketitle
\begin{abstract}
Data augmentation methods for neural machine translation are particularly useful when limited amount of training data is available, which is often the case when dealing with low-resource languages. We introduce a novel augmentation method, which generates new sentences by swapping objects and subjects across bisentences. This is performed simultaneously based on the dependency parse trees of the source and target sentences. We name this method \textit{TreeSwap}.
Our results show that TreeSwap achieves consistent improvements over baseline models in 4 language pairs in both directions on resource-constrained datasets. We also explore domain-specific corpora, but find that our method does not make significant improvements on law, medical and IT data. We report the scores of similar augmentation methods and find that TreeSwap performs comparably. We also analyze the generated sentences qualitatively and find that the augmentation produces a correct translation in most cases. Our code is available on Github\footnote{\url{https://github.com/attilanagy234/TreeSwap}, last accessed on 31/07/23}.
\end{abstract}

\section{Introduction}
Most Natural Language Processing (NLP) problems are formulated as supervised learning tasks, where large amounts of data is required to train models. Collecting annotated datasets is often time-consuming and laborious, so this motivated a lot of work in NLP to create methods for generating synthetic data that improves the dataset used for training in both size and variety, ultimately leading to more performant models \citep{feng-etal-2021-survey}. These Data Augmentation (DA) methods not only help in resource-constrained scenarios, but can also improve class imbalance \citep{chawla2002smote}, mitigate bias \citep{zhao-etal-2018-gender}, make the model more robust to out of distribution inputs \citep{yao2022improving} or simply improve model accuracy. An efficient data augmentation method for any NLP task has two main objectives, which need to be balanced: the augmented data should be diverse enough, that it provides new information during training, but it should also be label-preserving to avoid injecting unwanted noise into the model. In machine translation, this means that our aim is to generate diverse sentence pairs from existing data such that the parallelism holds.

In this paper, we propose TreeSwap, a data augmentation method for Neural Machine Translation (NMT) using dependency parsing. The core idea of TreeSwap is to find corresponding subtrees in the dependency parse trees of a translation pair and swap these to generate new data. As our augmentation procedure is based on dependency parsing with some additional rules to improve grammatical and morphological correctness, the generated sentence pairs are semantically nonsensical in many cases. Using such nonsensical or \emph{nonce}, but syntactically correct sentences as training data has been studied before and shown to perform well even when models cannot rely on semantic or lexical cues \citep{gulordava-etal-2018-colorless}. To demonstrate the effectiveness of TreeSwap, we perform resource-constrained experiments on 4 language pairs in both directions. We also train models on domain-specific corpora and evaluate on both in-domain and out-of-domain test sets. We compare our results to other common augmentation methods in NMT using standard machine translation metrics. To study the quality of the generated sentences and understand the possible errors in the augmentation, we also perform a qualitative analysis on the synthetic sentence pairs.

\section{Related Work}
In the context of machine translation, backtranslation \citep{sennrich-etal-2016-improving} has been the most dominant DA method. It uses monolingual data in the target language to generate new training samples. Backtranslation and its variants were shown to boost translation quality at multiple scales \citep{edunov-etal-2018-understanding} and demonstrate SOTA results on many language pairs \citep{hoang2018iterative}. \citet{fadaee-etal-2017-data} select rare words in the corpus and replace these in new contexts simultaneously in the source and target sentences. \citet{norouzi2016reward} introduce Reward Augmented Maximum Likelihood (RAML), which replaces words in the target sentence with other words from the target vocabulary. SwitchOut \citep{wang-etal-2018-switchout} is an extension of RAML, where the augmentation is performed on both the source and target sentences. Instead of selecting words from the vocabulary for replacement, SeqMix \citep{guo-etal-2020-sequence} randomly combines two sentences from the input. \citet{gao-etal-2019-soft} introduce Soft Contextual DA, where they replace the embedding of a random word with a weighted combination of other semantically similar, related words. \citet{duan2020syntax} use the depth of tokens in the dependency tree for weighting the selection probabilities of tokens for blanking, dropout and replacement. \citet{NEURIPS2020_7221e5c8} augment by merging the predictions of multiple forward and backward models with the original dataset. \citet{moussallem2019augmenting} improve the translation of entities and terminological expressions using knowledge graphs for augmentation. \citet{sanchez-cartagena-etal-2021-rethinking} apply simple transformations that are used as auxiliary tasks in a multi-task learning framework with the aim of providing new contexts during prediction. \citet{wei-etal-2022-learning} propose Continuous Semantic Augmentation (CSANMT), which augments each training instance with an adjacency semantic region to cover synonymous representations.

Syntax-based augmentation methods have been shown effective in a number of NLP tasks. \citet{xu-etal-2016-improved} use the directionality of relationships in a dependency tree to improve relation classification models. \citet{sahin-steedman-2018-data} generate augmented data for part-of-speech tagging by morphing the dependency tree through cropping edges and performing rotations around the root. \citet{vania-etal-2019-systematic} extend this method for dependency parsing and also apply another augmentation called nonce sentence generation, inspired by \citet{gulordava-etal-2018-colorless}. \citet{dehouck-gomez-rodriguez-2020-data} extends the subtree swapping method to augment data for dependency parsing. They perform the swapping in a more generic setting, not only on subjects and objects, but apply a wide range of morphological and structural constraints to ensure grammatical correctness. \citet{shi-etal-2020-role} see improvements in few-shot constituency parsing by dependency subtree substitution. \citet{shi-etal-2021-substructure} present a generalization of the previous methods and perform experiments on multiple NLP tasks. For reference, preliminary results of TreeSwap have been published prior to this paper \citep{nagy2023data}.

\section{Methodology}

\begin{figure*}[t!]
    \includegraphics[width=0.99\textwidth]{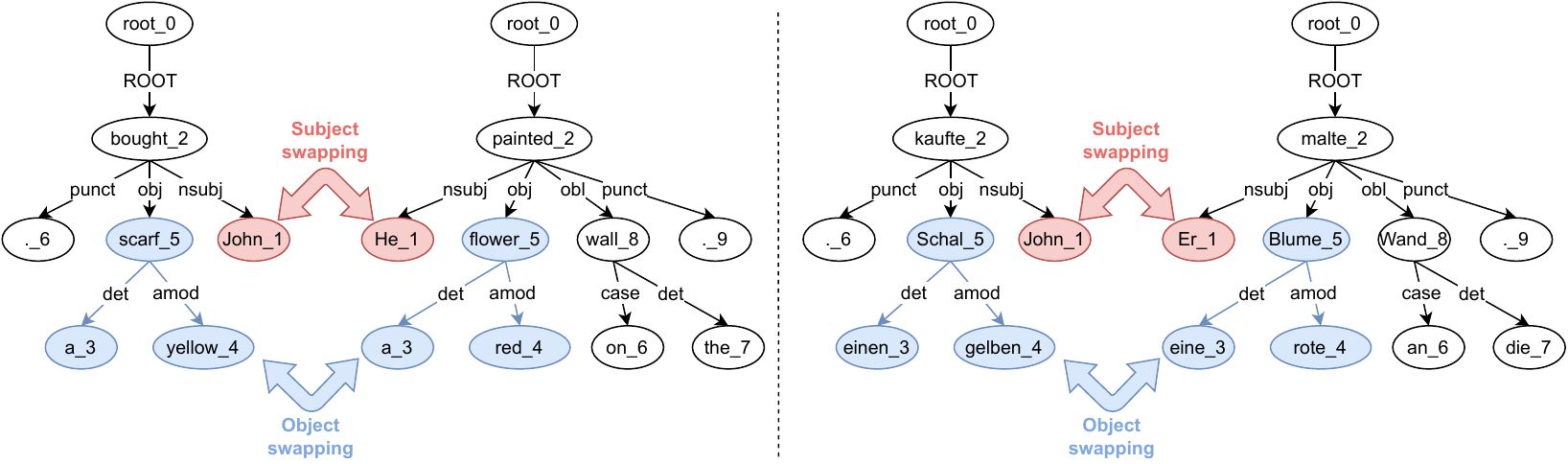}
    \caption{Two kinds of augmentation techniques: object and subject subtree swapping.}
    \label{fig:swapping}
\end{figure*}

\subsection{Subtree swapping}
Let $S = {s_1, s_2, \ldots, s_n}$ and $T = {t_1, t_2, \ldots, t_n}$ be a parallel corpus of source and target sentences, respectively. Our proposed data augmentation is based on the extraction of syntactic structures from the source and target sentences using dependency parsing. We denote the dependency parse of a sentence $s$ as $\text{Dep}(s)$, which is a directed graph $G = (V, E)$ representing the syntactic structure of $s$, where $V$ is the set of vertices representing words in $s$, and $E$ is the set of directed edges representing dependencies between words. We define a syntactic subtree of a sentence $s$ rooted at a word $v$ as the subgraph of $Dep(s)$ that includes $v$ and all its descendants. We denote the syntactic subtree rooted at $v$ as $ST(v, s)$.

Given two parallel sentence pairs $(s_1, t_1)$ and $(s_2, t_2)$, augmentation via subtree swapping can be defined as:
\begin{equation}
\begin{aligned}
    \label{eq:subtree_swapping}
    s_{\text{aug}} & = \text{replace}(\text{ST}(v, s_1), \text{ST}(u, s_2), s_1) \\ %\nonumber
    t_{\text{aug}} & = \text{replace}(\text{ST}(x, t_1), \text{ST}(y, t_2), t_1)
\end{aligned}
\end{equation}
where $\text{replace}(\text{ST}_1, \text{ST}_2, s)$ denotes the sentence obtained by replacing the syntactic subtree rooted at $\text{ST}_1$ in $s$ with the subtree rooted at $\text{ST}_2$, and $v$, $u$, $x$ and $y$ are subtree roots corresponding to the original sentence pair. To ensure that the resulting sentence pair remains a parallel translation, we apply a number of constraints on the algorithm. 

\begin{itemize}
  \item We only extract two types of subtrees from sentences: objects and subjects. We consider these subtrees to correspond to the $\text{OBJ}$ and $\text{NSUBJ}$ dependency edges defined in the Universal Dependencies \citep{nivre-etal-2020-universal}. We experimented with extracting more complex substructures such as predicates for subtree swapping, but found that it did not generalize well across language-pairs and likely injected too much noise into the training data via augmentation.
  \item The dependency trees of both the source and target sentences must contain exactly one $\text{OBJ}$ and $\text{NSUBJ}$ edge.
  \item The source and target subtree roots must belong to the same part of speech tag.
  \item Every selected subtree must contain at least a noun or a proper noun
\end{itemize}

 The method is illustrated in Figure \ref{fig:swapping}.

\subsection{Sampling}
As the pair-wise subtree swaps can produce a quadratically large number of augmented sentences with respect to original data, we experiment with two sampling methods alongside a random sampling baseline. The key to both methods is observing the syntactic structure of the extracted subtrees in the dependency parse tree. We apply two graph similarity metrics on the subtrees and use this as a bias for sampling later in our experiments.
\paragraph{Graph Edit Distance (GED)} 
Similar to the Levenshtein distance \citep{levenshtein1966binary}, GED \citep{6313167} defines the minimal number of operations (insertion, deletion and substitution) required to transform a graph into another. The weight of deletions and insertions is 1, for substitution it is 2. To make sure GED is comparable regardless of graph size, we normalize it as such:
\begin{equation}
\begin{aligned}
    \label{eq:ged}
    d_{\text{max}} &= 2 |V_1| - 1 + 2 |V_2| - 1 \\ 
    \text{sim}(G_1,G_2) &= \dfrac{d_{\text{max}} - \text{GED}(G_1,G_2)}{d_{\text{max}}} %\nonumber 
\end{aligned}
\end{equation}
where $d_{\text{max}}$ is the maximum distance between two graphs.

\paragraph{Edge Mapping (EM)}

EM is based on the labeled graph similarity measure of \citet{graph_sim}. A $\text{score}(e_1, e_2)$ function denotes the number of common nodes between two edges. Given two edges $e_1$ and $e_2$, we take the routes in the graph from the root to $e_1$ and $e_2$ respectively and define the route by the part of speech tags of the nodes that are visited from the root to the edges. The $\text{route\_sim}(e_1, e_2)$ function computes the Levenshtein distance between two such routes. With the help of Algorithm~\ref{alg:mapping}, we can compute a mapping between the edges of the graph. Using this mapping, we can calculate a Jaccard index between the edges, which now can serve as a similarity measure between the dependency trees:

\begin{equation}
\label{eq:jaccard}
    J(G_1,G_2) = \dfrac{|m|}{|E_1| + |E_2| - |m|} %\nonumber
\end{equation}
where $m$ is the mapping, $E_1$ and $E_2$ are the set of edges in $G_1$ and $G_2$ respectively.

\begin{algorithm}[t]
\caption{Edge mapping.}
\label{alg:mapping}
\begin{algorithmic}
\Require $G_1(V_1,E_1),G_2(V_2,E_2)$
\State $\text{mapping} \gets \{\}$
\ForAll {$e_1 \in E_1$}
\State $\text{cands} \gets \{e_2 \mid e_2 \in E_2, e_1 \neq e_2,  e_2 \notin \text{mapping}\}$
\If{$\text{cands}$ is empty}
    \State \textbf{continue}
\EndIf
\State $\text{cands} \gets \argmax\limits_{c \in \text{cands}} \text{score}(e_1, c)$
\State $\text{cands} \gets \argmax\limits_{c \in \text{cands}} \text{route\_sim}(e_1, c)$
\State $\text{mapping}[e_1] \gets \text{random}(\text{cands})$
\EndFor
\State \Return mapping
\end{algorithmic}
\end{algorithm}

\section{Experiments}
We conduct experiments on 4 language pairs, English to German, Hebrew, Vietnamese and Hungarian in both directions. We selected corpora that are considered low-resource and widely used in the community to evaluate data augmentation approaches for machine translation. We also perform domain-specific experiments in three domains, evaluating the effectiveness of the DA method on both in-domain and out-of-domain setups. We ran all experiments 3 times with different seeds for robust results.

\paragraph{Datasets}
For English-German and English-Hebrew we use the IWSLT 2014 text translation track \citep{cettolo-etal-2014-report} datasets for training data as done by \citet{gao-etal-2019-soft}, \citet{guo-etal-2020-sequence} and \citet{sanchez-cartagena-etal-2021-rethinking}. For development and testing we use the \textit{tst2013} and \textit{tst2014} datasets. For English-Vietnamese, we use the IWSLT 2015 text translation track \citep{cettolo-etal-2015-iwslt} dataset with the \textit{tst2012} and \textit{tst2013} datasets used for development and testing as done by \citet{wang-etal-2018-switchout} and \citet{sanchez-cartagena-etal-2021-rethinking}. For Hungarian-English, we produce a subsample comparable in size to the IWSLT datasets using the Hunglish2 corpus \citep{varga2007parallel}. As low-resource datasets are usually composed of a few sources and they generally are not linguistically diverse, we decided to only sample from the modern literature subcorpus of Hunglish2 and discard the others. This should still be considered as a high-resource experiment with withheld data, although we try to mimic a low-resource scenario as much as possible. Following \citet{wang-sennrich-2020-exposure} and \citet{sanchez-cartagena-etal-2021-rethinking}, we use the IT, law and medical domain-specific datasets published by \citep{muller-etal-2020-domain}. The statistics of the datasets are summarized in Table \ref{table:dataset-stats}.

\begin{table}[!htbp]
  \centering
    \begin{tabular}{ l r r r }
    \toprule
    \textbf{Dataset} & \textbf{train} & \textbf{dev} & \textbf{test} \\
    \midrule
    En-De & 174,443 & 993 & 1,305 \\
    En-He & 187,817 & 1,382 & 962 \\
    En-Vi & 133,317 & 1,553 & 1,268 \\
    En-Hu & 120,000 & 2,000 & 2,000 \\
    \midrule
    IT & 265,179 & 2,000 & 2,000 \\
    Law & 501,379 & 2,000 & 2,000 \\
    Medical & 360,249 & 2,000 & 2,000 \\
    \bottomrule
    \end{tabular}
  \caption{Number of bisentences in the preprocessed train/dev/test sets for each language pair and domain.}
  \label{table:dataset-stats}
\end{table}

\paragraph{Preprocessing}
In the English-German, English-Hebrew and English-Vietnamese IWSLT experiments we decided to use the same preprocessing steps as \citet{sanchez-cartagena-etal-2021-rethinking} and we also use their train, development, and test splits for comparable results. For English-Hungarian we remove sentences if they are longer than 32 tokens or if the source-target token count difference is more than 7 and their ratio is more than 1.2. We also strip leading and trailing quotation marks and dashes and normalize punctuations with \textit{sacremoses}\footnote{\url{https://github.com/alvations/sacremoses}, last accessed on 31/07/23}. We also infer the source and target languages with \textit{fastText} \citep{joulin2016fasttext} and remove sentence pairs in case of a mismatch. For the English-German domain specific corpora, we use a maximum word count of 100 and a maximum word count difference of 10 between the source and target sentences. We also removed duplicated sentence pairs from the data and created a new train/dev/test split. Overall, the deduplication considerably reduced the size of the datasets in all three domains.

\begin{figure*}[!htbp]
    \includegraphics[width=0.99\textwidth]{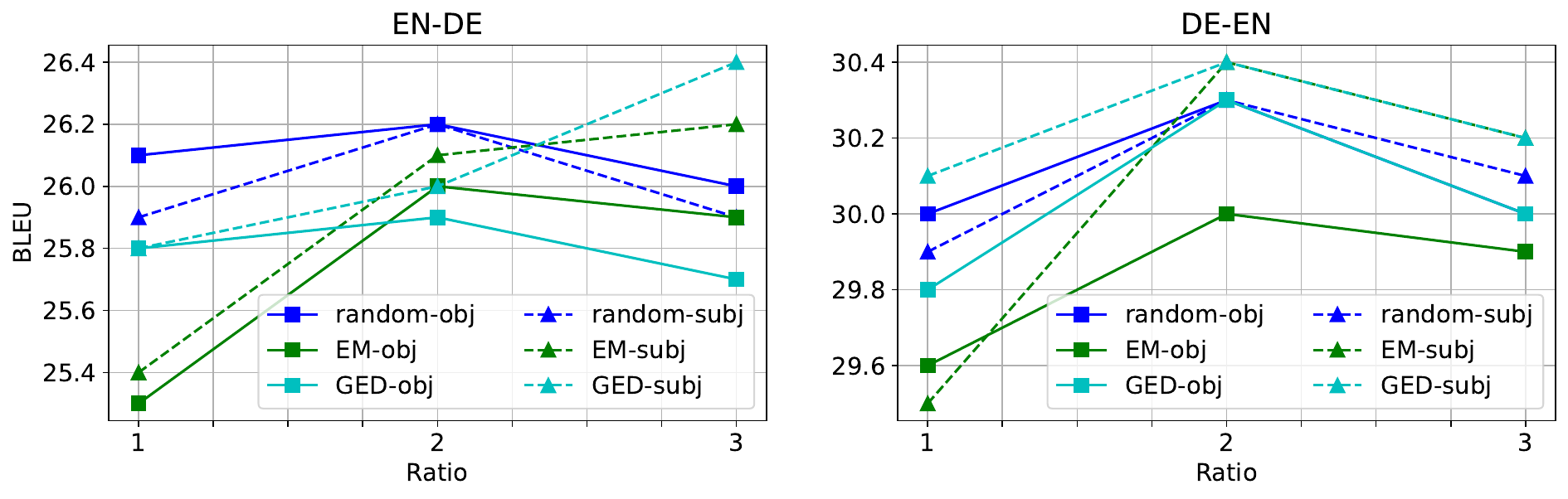}
    \caption{The results of tuning the sampling method parameter for the English-German language pair.}
    \label{fig:sampling_tuning}
\end{figure*}

\begin{figure}[H]
    \includegraphics[width=0.99\linewidth]{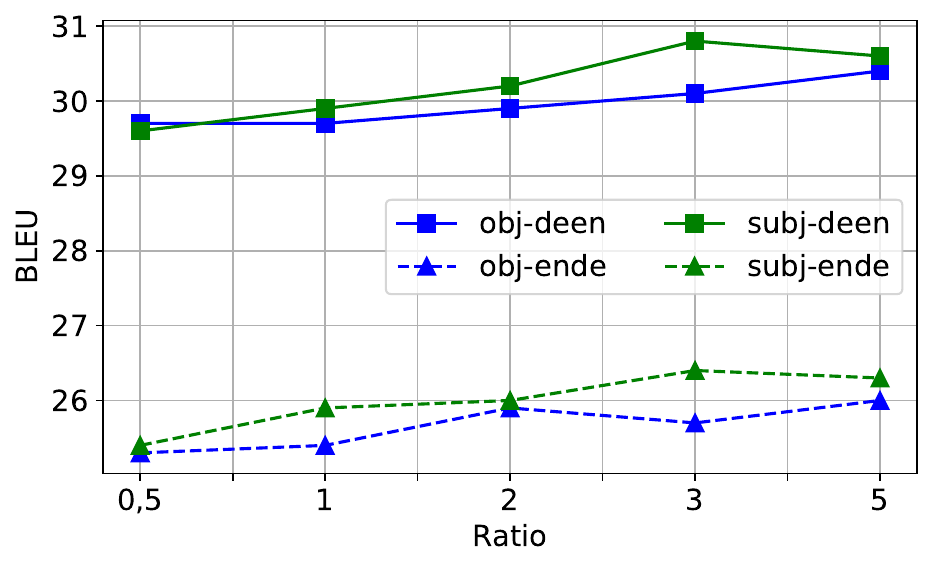}
    \caption{The BLEU scores of experimenting with different ratios.}
    \label{fig:ratio_tuning}
\end{figure}

\paragraph{Augmentation details}
In all of our experiments, we only mix augmented data into the training sets, while development and test sets are left untouched. Due to the vast number of combinations resulting from our augmentation method's multiple hyperparameters, we decide to tune every parameter individually. The first one is the sampling threshold that we measure for every language pair separately. We find that 0.5 works for every pair the best.
Moving forward, we only do experiments on the English-German pair, due to computational limits. The next parameter is the sampling method, we run experiments with ratios 1, 2 and 3 in both directions. According to the BLEU scores that are presented in Figure \ref{fig:sampling_tuning}, we choose the GED method for further experiments. Next, we study the augmentation ratio only with the GED method with 0.5 similarity threshold. Figure \ref{fig:ratio_tuning} shows the BLEU scores of our experiments. We decide to do every further augmentation with the GED sampling method, using 0.5 threshold and 3 as the augmentation ratio. For dependency parsing we use huspacy \citep{HuSpaCy:2021} for Hungarian and Stanza \citep{qi2020stanza} for every other language.

\paragraph{Training details}
We train the same encoder-decoder model for every language pair based on the Transformer architecture \citep{vaswani2017attention}. All hyperparameters of the model can be found in Table \ref{table:base-parameters}. The models were implemented in Python using the openNMT framework \citep{klein-etal-2017-opennmt}. Every model was trained with early stopping to avoid overfitting, using the validation perplexity as a stopping criterion. The training jobs were executed on a cluster of machines with A100 GPUs.

\begin{table*}[!htbp]
\centering
\begin{tabular}{lr|lr}
\toprule
\textbf{Parameter}          & \textbf{Value}    & \textbf{Parameter}    & \textbf{Value}       \\ \midrule
batch type & tokens & batch size & 3000-8000\\
accumulation count & 4 & average decay & 0.0005 \\
train steps & 150000 & valid steps & 5000 \\
early stopping & 4 & early stopping criteria & ppl \\
optimizer & adam & learning rate & 2 \\
warmup steps & 8000 & decay method & noam \\
adam beta2 & 0.998 & max grad norm & 2 \\
label smoothing & 0.1 & param init & 0 \\
param init glorot & true & normalization & tokens \\
max generator batches & 32 & encoder layers & 8 \\
decoder layers & 8 & heads & 16 \\
RNN size & 1024 & word vector size & 1024 \\
Transformer FF & 2096 & dropout steps & 0\\
dropout & 0.1 & attention dropout & 0.1\\
share embeddings & true & position encoding & true \\
\bottomrule
\end{tabular}
\caption{Hyperparameters of the models.}
\label{table:base-parameters}
\end{table*}

\section{Results}

In order to measure the effectiveness of TreeSwap, we used common evaluation metrics such as the BLEU and the METEOR scores. These scores were computed for both the augmented and baseline models to enable a comparative analysis of the proposed method against previous augmentation approaches. The results of these analyses are presented in Table \ref{table:results} and Table \ref{table:iwslt}.

\subsection{Quantitative evaluation}

Our results demonstrate that each of the examined DA methods consistently improves translation quality across all language pairs. Specifically, Table \ref{table:results} showcases that the subject-based approach consistently outperforms other augmentation strategies, leading to a substantial increase in BLEU scores by 0.5-1 points. Further, our findings indicate that the subject based DA technique yields the most favorable outcomes based on METEOR scores, with an improvement of 0.5-1 points.

We also compared the effectiveness of our DA techniques with previous augmentation methods. The results demonstrate that the TreeSwap augmentation method consistently outperforms SwitchOut+RAML and approaches the results of reverse+mono+replace, even outperforming the latter in the case of Vietnamese-English. These results confirm that the TreeSwap technique holds great promise as a reliable augmentation strategy to enhance the performance of NMT systems.

Table \ref{table:domain_results} represents the results of our in-domain and out-of-domain experiments. The TreeSwap augmentation did not yield any significant improvements in translation for domain-specific datasets. Our baseline reached the highest scores in both the in-domain and the out-of-domain experiments.

\begin{table*}
\centering
\begin{tabular}{lrrrrrr}
\toprule
{} & \multicolumn{3}{c}{BLEU} & \multicolumn{3}{c}{METEOR} \\
{} &       base &        object &       subject &     base &      object &     subject \\
\midrule
\midrule
de-en &   29.60$\pm$0.1 &  30.03$\pm$0.1 &  \textbf{30.37$\pm$0.2} &   60.7$\pm$0.1 &  61.07$\pm$0.1 &  \textbf{61.31$\pm$0.1} \\
en-de &   25.60$\pm$0.5 &  \textbf{26.17$\pm$0.3} &  \textbf{26.17$\pm$0.2} &  53.92$\pm$0.2 &  54.25$\pm$0.1 &  \textbf{54.38$\pm$0.1} \\
he-en &  31.43$\pm$0.3 &  32.13$\pm$0.3 &  \textbf{32.53$\pm$0.2} &  63.25$\pm$0.1 &  63.71$\pm$0.3 &  \textbf{64.03$\pm$0.3} \\
en-he &   21.40$\pm$0.3 &  21.93$\pm$0.3 &  \textbf{22.03$\pm$0.3} &  47.54$\pm$0.2 &  48.19$\pm$0.3 &  \textbf{48.21$\pm$0.3} \\
vi-en &  29.77$\pm$0.2 &  \textbf{29.97$\pm$0.2} &  29.73$\pm$0.3 &  59.54$\pm$0.2 &  \textbf{59.55$\pm$0.3} &  \textbf{59.55$\pm$0.1} \\
en-vi &   29.20$\pm$0.0 &   29.5$\pm$0.3 &  \textbf{29.77$\pm$0.3} &  58.86$\pm$0.0 &  58.63$\pm$0.4 &  \textbf{59.04$\pm$0.3} \\
hu-en &  10.63$\pm$0.2 &  \textbf{11.93$\pm$0.1} &  11.83$\pm$0.2 &   34.9$\pm$0.2 &   \textbf{36.6$\pm$0.3} &  36.46$\pm$0.2 \\
en-hu &   8.03$\pm$0.1 &   8.47$\pm$0.2 &   \textbf{8.83$\pm$0.2} &  30.58$\pm$0.1 &  31.07$\pm$0.2 &  \textbf{31.54$\pm$0.3} \\
\bottomrule
\end{tabular}
\caption{BLEU and METEOR scores of the IWSLT and hu-en experiments.}
  \label{table:results}
\end{table*}

\begin{table*}
\centering
\begin{tabular}{lcccccc}
\toprule
      &          en-de &          de-en &          en-he &          he-en &          en-vi &          vi-en \\
\midrule
\midrule
their baseline &   24.7$\pm$0.2 &   30.0$\pm$0.1 &   21.5$\pm$0.3 &  32.4$\pm$0.1 &   28.9$\pm$0.1 &  27.5$\pm$0.4 \\
our baseline  &   25.6$\pm$0.5 &   29.6$\pm$0.1 &   21.4$\pm$0.3 &  31.4$\pm$0.3 &   29.2$\pm$0.0 &  29.8$\pm$0.2 \\
\midrule

SwitchOut & 25.3$\pm$0.2 &  30.1$\pm$0.2 &  21.6$\pm$0.6 &  32.1$\pm$0.4 &  28.5$\pm$0.2 &  27.3$\pm$0.6 \\

RAML & 25.4$\pm$0.2 &  30.3$\pm$0.1 &  21.9$\pm$0.1 &  32.1$\pm$0.1 &  28.6$\pm$0.5 &  27.3$\pm$0.5 \\

SwitchOut+RAML & 25.7$\pm$0.4 &  30.3$\pm$0.5 &  22.1$\pm$0.4 &  32.1$\pm$0.4 &  29.1$\pm$0.4 &  27.5$\pm$0.3 \\

reverse+mono+replace  &  26.4$\pm$0.6 &  31.4$\pm$0.3 &  23.2$\pm$0.3 &  33.9$\pm$0.5 &  30.5$\pm$0.2 &  29.4$\pm$0.3 \\

\midrule
TreeSwap  &  26.2$\pm$0.2 &  30.4$\pm$0.2 &  22.0$\pm$0.3 &  32.5$\pm$0.2 &  29.8$\pm$0.3 &  30.0$\pm$0.2 \\

\bottomrule
\end{tabular}
\caption{Comparison of TreeSwap to other augmentation methods for NMT. The reported scores are based on the implementations of \citet{sanchez-cartagena-etal-2021-rethinking}.}
  \label{table:iwslt}
\end{table*}

\begin{table*}
\begin{tabular}{llrrrrrr}
\toprule
 & & \multicolumn{3}{c}{de-en} & \multicolumn{3}{c}{en-de} \\
      train        & test &             baseline &            object &           subject &  baseline &            object &           subject \\
\midrule
\midrule
 it & it &  37.60$\pm$0.8 &  37.57$\pm$0.1 &  36.60$\pm$0.8 &  32.97$\pm$0.2 &  32.83$\pm$0.1 &  32.37$\pm$0.6 \\
    & law &   5.57$\pm$0.3 &   4.77$\pm$0.5 &   5.23$\pm$0.2 &   4.93$\pm$0.4 &   4.07$\pm$0.1 &   4.13$\pm$0.3 \\
    & medical &   5.83$\pm$0.4 &   4.83$\pm$0.4 &   4.93$\pm$0.2 &   5.17$\pm$0.2 &   3.83$\pm$0.2 &   4.20$\pm$0.4 \\
\midrule
law & it &   4.87$\pm$0.7 &   4.80$\pm$0.2 &   4.57$\pm$0.5 &   3.67$\pm$0.1 &   4.10$\pm$0.6 &   4.37$\pm$0.5 \\
    & law &  59.53$\pm$0.3 &  58.77$\pm$0.4 &  58.93$\pm$0.6 &  54.07$\pm$0.2 &  53.20$\pm$0.1 &  53.33$\pm$0.2  \\
    & medical &   9.33$\pm$0.3 &   9.10$\pm$0.1 &   8.47$\pm$0.5  & 8.83$\pm$0.7 &   8.77$\pm$0.2 &   8.37$\pm$0.2  \\
\midrule
medical & it &   2.80$\pm$0.3 &   2.37$\pm$0.4 &   2.27$\pm$0.4  &   2.23$\pm$0.3 &   2.07$\pm$0.3 &   2.23$\pm$0.2 \\
        & law &   7.90$\pm$0.5 &   6.67$\pm$0.6 &   6.30$\pm$0.4 &   5.77$\pm$0.2 &   5.07$\pm$0.5 &   5.00$\pm$0.6  \\
        & medical &  56.97$\pm$0.5 &  55.90$\pm$1.0 &  56.47$\pm$0.4 &  52.67$\pm$0.3 &  51.70$\pm$0.3 &  51.80$\pm$0.6  \\
\bottomrule
\end{tabular}
\caption{The BLEU scores for the in-domain and the out-of-domain experiments.}
\label{table:domain_results}
\end{table*}

\subsection{Qualitative evaluation}
Improvements in automated metrics such as BLEU or METEOR give some idea about the effectiveness of an augmentation method, but they do not provide insights into the quality of the generated sentences. To better understand the behaviour of TreeSwap, we run a qualitative analysis on a small sample of English-German translations, including both augmented and original data. We hired 3 annotators, who possess at least a B2 level certification in both English and German. We asked them to assess the quality of 150 sentence pairs sampled from the EN-DE IWSLT train set. Out of the 150 sentence pairs, 50 were original data points without augmentation, 50 were generated via subject swapping and 50 via object swapping. Apart from the sentences, the annotators could view the the parts of the sentences that are extracted for augmentation. The annotators had to answer the following questions:
\begin{itemize}
    \item \textbf{Question A:} Is the sentence pair a correct translation?
    \item \textbf{Question B:} Is the English sentence grammatically correct?
    \item \textbf{Question C:} Is the German sentence grammatically correct?
    \item \textbf{Question D:} Do the extracted parts in the source and target sentences correspond to the same meaning?
\end{itemize}
With \textit{Question A} our intention is to get an idea about the quality of translations in general and what portion of the generated data can be considered useful for training. As our method does not adapt the morphology or grammar of the swapped subtrees, we explore the extent of this with \textit{Question B} and \textit{Question C}. If the subtrees in the source and target sentences that are extracted for swapping do not mean the same thing, the augmentation is very likely to violate the parallelism of the translation pair. We measure this with \textit{Question D}.

\begin{figure}[htb!]
    \includegraphics[width=0.99\linewidth]{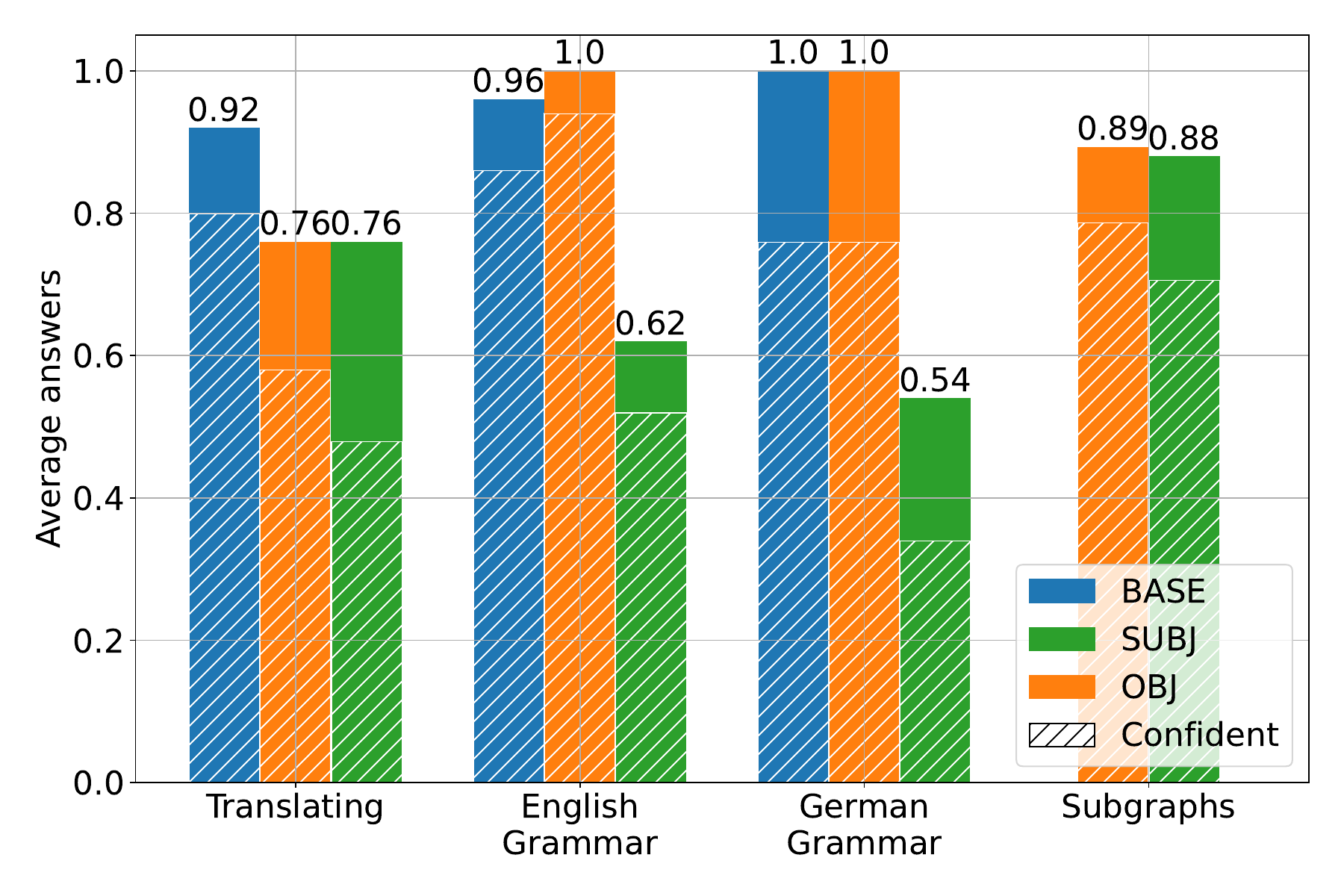}
    \caption{Results of the qualitative evaluation. The proportion of confident annotations (all three annotators agreed) are highlighted.}
    \label{fig:qualitative_results}
\end{figure}

The results of the evaluation are summarized in Figure \ref{fig:qualitative_results}. The quality of the augmented sentences turned out to be equally good for the subject and object swapping with 76\% of the sentence pairs considered as correct translations. The annotators were instructed that a translation can be considered correct with a minor grammatical mistake. The grammatical correctness of the base sentences and the object swapping augmented sentences is on par, while the subject subtree swapping resulted in a significantly higher number of errors. We observed during our experiments and also received feedback from the annotators that sentences were often problematic when personal pronouns are swapped as the subject subtree, since inflection in the sentence is dependent on the pronouns. Interestingly, despite the high number of grammatical errors, subject swapping seemed to produce the highest BLEU scores on most language pairs. We also saw high correlation between answers to question \textit{A} and \textit{D}, having different answers only in 15.3\% of the cases. There were only 6 cases, where the extracted subgraphs were identified as having a different meaning, but the augmented sentence pair was marked as a correct translation. This indicates that the performance of the augmentation is largely dependent on the quality of the underlying dependency parser. We compute Cohen's kappa \citep{cohen_kappa} to measure inter-annotator agreement. The average pair-wise Cohen's kappa was 49.6\% indicating moderate agreement. The translation correctness had the lowest Cohen's kappa with 41.1\%. For the grammatical correctness questions, the annotators showed more agreement for English with 69.9\%, compared to a kappa of 43.5\% for German. The question about whether the extracted subgraphs match had a Cohen's kappa of 46.3\%.

\section{Conclusion}
In this paper we presented a new data augmentation method for NMT that we call TreeSwap. Our method generates new samples by swapping compatible subtrees of the dependency parse trees of translation-pairs. More precisely we swap objects and subjects simultaneously in the source and target sentences between two translation pairs to generate new parallel translations. Experiments on 4 language pairs in both directions have shown that models trained with data augmented using TreeSwap can consistently outperform baseline models. We also compared TreeSwap to other augmentation methods used in NMT and found that TreeSwap achieves compatible performance to other methods. However, with domain-specific corpora, TreeSwap brought little to no performance gains in terms of quantitative metrics, which suggests that the type of corpora used for augmentation heavily influences the success of our method. Our qualitative analysis has shown that the generated sentences are pre-dominantly correct translations, but also revealed that TreeSwap can induce certain undesired grammatical errors. It is an interesting future direction to explore how these issues could be fixed either via heuristics or fixing the morphosyntactic errors with another model. The improvements by TreeSwap (like many other augmentation methods) seem to depend on finding a good balance between distorting the translation distribution and enriching the model with synthetic translation pairs. It would be interesting to study the change in translation distributions induced by TreeSwap.

\section*{Acknowledgments}
The authors would like to thank András Kornai and Csaba Oravecz for discussions on TreeSwap.

%The acknowledgments should go immediately before the references. Do not number the acknowledgments section.
%\textbf{Do not include this section when submitting your paper for review.}
 
\bibliographystyle{acl_natbib}
\bibliography{ranlp2023}

\begin{thebibliography}{40}
\expandafter\ifx\csname natexlab\endcsname\relax\def\natexlab#1{#1}\fi

\bibitem[{Cettolo et~al.(2015)Cettolo, Niehues, St{\"u}ker, Bentivogli,
  Cattoni, and Federico}]{cettolo-etal-2015-iwslt}
Mauro Cettolo, Jan Niehues, Sebastian St{\"u}ker, Luisa Bentivogli, Roldano
  Cattoni, and Marcello Federico. 2015.
\newblock \href {https://aclanthology.org/2015.iwslt-evaluation.1} {The {IWSLT}
  2015 evaluation campaign}.
\newblock In \emph{Proceedings of the 12th International Workshop on Spoken
  Language Translation: Evaluation Campaign}, pages 2--14, Da Nang, Vietnam.

\bibitem[{Cettolo et~al.(2014)Cettolo, Niehues, St{\"u}ker, Bentivogli, and
  Federico}]{cettolo-etal-2014-report}
Mauro Cettolo, Jan Niehues, Sebastian St{\"u}ker, Luisa Bentivogli, and
  Marcello Federico. 2014.
\newblock \href {https://aclanthology.org/2014.iwslt-evaluation.1} {Report on
  the 11th {IWSLT} evaluation campaign}.
\newblock In \emph{Proceedings of the 11th International Workshop on Spoken
  Language Translation: Evaluation Campaign}, pages 2--17, Lake Tahoe,
  California.

\bibitem[{Champin and Solnon(2003)}]{graph_sim}
Pierre-Antoine Champin and Christine Solnon. 2003.
\newblock \href {https://doi.org/10.1007/3-540-45006-8_9} {Measuring the
  similarity of labeled graphs}.
\newblock volume 2689.

\bibitem[{Chawla et~al.(2002)Chawla, Bowyer, Hall, and
  Kegelmeyer}]{chawla2002smote}
Nitesh~V Chawla, Kevin~W Bowyer, Lawrence~O Hall, and W~Philip Kegelmeyer.
  2002.
\newblock Smote: synthetic minority over-sampling technique.
\newblock \emph{Journal of artificial intelligence research}, 16:321--357.

\bibitem[{Cohen(1960)}]{cohen_kappa}
Jacob Cohen. 1960.
\newblock \href {https://doi.org/10.1177/001316446002000104} {A coefficient of
  agreement for nominal scales}.
\newblock \emph{Educational and Psychological Measurement}, 20(1):37--46.

\bibitem[{Dehouck and
  G{\'o}mez-Rodr{\'\i}guez(2020)}]{dehouck-gomez-rodriguez-2020-data}
Mathieu Dehouck and Carlos G{\'o}mez-Rodr{\'\i}guez. 2020.
\newblock \href {https://doi.org/10.18653/v1/2020.coling-main.339} {Data
  augmentation via subtree swapping for dependency parsing of low-resource
  languages}.
\newblock In \emph{Proceedings of the 28th International Conference on
  Computational Linguistics}, pages 3818--3830, Barcelona, Spain (Online).
  International Committee on Computational Linguistics.

\bibitem[{Duan et~al.(2020)Duan, Zhao, Zhang, and Wang}]{duan2020syntax}
Sufeng Duan, Hai Zhao, Dongdong Zhang, and Rui Wang. 2020.
\newblock Syntax-aware data augmentation for neural machine translation.
\newblock \emph{arXiv preprint arXiv:2004.14200}.

\bibitem[{Edunov et~al.(2018)Edunov, Ott, Auli, and
  Grangier}]{edunov-etal-2018-understanding}
Sergey Edunov, Myle Ott, Michael Auli, and David Grangier. 2018.
\newblock \href {https://doi.org/10.18653/v1/D18-1045} {Understanding
  back-translation at scale}.
\newblock In \emph{Proceedings of the 2018 Conference on Empirical Methods in
  Natural Language Processing}, pages 489--500, Brussels, Belgium. Association
  for Computational Linguistics.

\bibitem[{Fadaee et~al.(2017)Fadaee, Bisazza, and Monz}]{fadaee-etal-2017-data}
Marzieh Fadaee, Arianna Bisazza, and Christof Monz. 2017.
\newblock \href {https://doi.org/10.18653/v1/P17-2090} {Data augmentation for
  low-resource neural machine translation}.
\newblock In \emph{Proceedings of the 55th Annual Meeting of the Association
  for Computational Linguistics (Volume 2: Short Papers)}, pages 567--573,
  Vancouver, Canada. Association for Computational Linguistics.

\bibitem[{Feng et~al.(2021)Feng, Gangal, Wei, Chandar, Vosoughi, Mitamura, and
  Hovy}]{feng-etal-2021-survey}
Steven~Y. Feng, Varun Gangal, Jason Wei, Sarath Chandar, Soroush Vosoughi,
  Teruko Mitamura, and Eduard Hovy. 2021.
\newblock \href {https://doi.org/10.18653/v1/2021.findings-acl.84} {A survey of
  data augmentation approaches for {NLP}}.
\newblock In \emph{Findings of the Association for Computational Linguistics:
  ACL-IJCNLP 2021}, pages 968--988, Online. Association for Computational
  Linguistics.

\bibitem[{Gao et~al.(2019)Gao, Zhu, Wu, Xia, Qin, Cheng, Zhou, and
  Liu}]{gao-etal-2019-soft}
Fei Gao, Jinhua Zhu, Lijun Wu, Yingce Xia, Tao Qin, Xueqi Cheng, Wengang Zhou,
  and Tie-Yan Liu. 2019.
\newblock \href {https://doi.org/10.18653/v1/P19-1555} {Soft contextual data
  augmentation for neural machine translation}.
\newblock In \emph{Proceedings of the 57th Annual Meeting of the Association
  for Computational Linguistics}, pages 5539--5544, Florence, Italy.
  Association for Computational Linguistics.

\bibitem[{Gulordava et~al.(2018)Gulordava, Bojanowski, Grave, Linzen, and
  Baroni}]{gulordava-etal-2018-colorless}
Kristina Gulordava, Piotr Bojanowski, Edouard Grave, Tal Linzen, and Marco
  Baroni. 2018.
\newblock \href {https://doi.org/10.18653/v1/N18-1108} {Colorless green
  recurrent networks dream hierarchically}.
\newblock In \emph{Proceedings of the 2018 Conference of the North {A}merican
  Chapter of the Association for Computational Linguistics: Human Language
  Technologies, Volume 1 (Long Papers)}, pages 1195--1205, New Orleans,
  Louisiana. Association for Computational Linguistics.

\bibitem[{Guo et~al.(2020)Guo, Kim, and Rush}]{guo-etal-2020-sequence}
Demi Guo, Yoon Kim, and Alexander Rush. 2020.
\newblock \href {https://doi.org/10.18653/v1/2020.emnlp-main.447}
  {Sequence-level mixed sample data augmentation}.
\newblock In \emph{Proceedings of the 2020 Conference on Empirical Methods in
  Natural Language Processing (EMNLP)}, pages 5547--5552, Online. Association
  for Computational Linguistics.

\bibitem[{Hoang et~al.(2018)Hoang, Koehn, Haffari, and
  Cohn}]{hoang2018iterative}
Vu~Cong~Duy Hoang, Philipp Koehn, Gholamreza Haffari, and Trevor Cohn. 2018.
\newblock Iterative back-translation for neural machine translation.
\newblock In \emph{Proceedings of the 2nd workshop on neural machine
  translation and generation}, pages 18--24.

\bibitem[{Joulin et~al.(2016)Joulin, Grave, Bojanowski, Douze, J{\'e}gou, and
  Mikolov}]{joulin2016fasttext}
Armand Joulin, Edouard Grave, Piotr Bojanowski, Matthijs Douze, H{\'e}rve
  J{\'e}gou, and Tomas Mikolov. 2016.
\newblock Fasttext. zip: Compressing text classification models.
\newblock \emph{arXiv preprint arXiv:1612.03651}.

\bibitem[{Klein et~al.(2017)Klein, Kim, Deng, Senellart, and
  Rush}]{klein-etal-2017-opennmt}
Guillaume Klein, Yoon Kim, Yuntian Deng, Jean Senellart, and Alexander Rush.
  2017.
\newblock \href {https://www.aclweb.org/anthology/P17-4012} {{O}pen{NMT}:
  Open-source toolkit for neural machine translation}.
\newblock In \emph{Proceedings of {ACL} 2017, System Demonstrations}, pages
  67--72, Vancouver, Canada. Association for Computational Linguistics.

\bibitem[{Levenshtein et~al.(1966)}]{levenshtein1966binary}
Vladimir~I Levenshtein et~al. 1966.
\newblock Binary codes capable of correcting deletions, insertions, and
  reversals.
\newblock In \emph{Soviet physics doklady}, volume~10, pages 707--710. Soviet
  Union.

\bibitem[{Moussallem et~al.(2019)Moussallem, Ar{\v{c}}an, Ngomo, and
  Buitelaar}]{moussallem2019augmenting}
Diego Moussallem, Mihael Ar{\v{c}}an, Axel-Cyrille~Ngonga Ngomo, and Paul
  Buitelaar. 2019.
\newblock Augmenting neural machine translation with knowledge graphs.
\newblock \emph{arXiv preprint arXiv:1902.08816}.

\bibitem[{M{\"u}ller et~al.(2020)M{\"u}ller, Rios, and
  Sennrich}]{muller-etal-2020-domain}
Mathias M{\"u}ller, Annette Rios, and Rico Sennrich. 2020.
\newblock \href {https://aclanthology.org/2020.amta-research.14} {Domain
  robustness in neural machine translation}.
\newblock In \emph{Proceedings of the 14th Conference of the Association for
  Machine Translation in the Americas (Volume 1: Research Track)}, pages
  151--164, Virtual. Association for Machine Translation in the Americas.

\bibitem[{Nagy et~al.(2023)Nagy, Lakatos, Barta, Nanys, and
  {\'A}cs}]{nagy2023data}
Attila Nagy, Dorina~Petra Lakatos, Botond Barta, Patrick Nanys, and Judit
  {\'A}cs. 2023.
\newblock Data augmentation for machine translation via dependency subtree
  swapping.
\newblock \emph{arXiv preprint arXiv:2307.07025}.

\bibitem[{Nguyen et~al.(2020)Nguyen, Joty, Wu, and Aw}]{NEURIPS2020_7221e5c8}
Xuan-Phi Nguyen, Shafiq Joty, Kui Wu, and Ai~Ti Aw. 2020.
\newblock Data diversification: A simple strategy for neural machine
  translation.
\newblock In \emph{Advances in Neural Information Processing Systems},
  volume~33, pages 10018--10029. Curran Associates, Inc.

\bibitem[{Nivre et~al.(2020)Nivre, de~Marneffe, Ginter, Haji{\v{c}}, Manning,
  Pyysalo, Schuster, Tyers, and Zeman}]{nivre-etal-2020-universal}
Joakim Nivre, Marie-Catherine de~Marneffe, Filip Ginter, Jan Haji{\v{c}},
  Christopher~D. Manning, Sampo Pyysalo, Sebastian Schuster, Francis Tyers, and
  Daniel Zeman. 2020.
\newblock \href {https://aclanthology.org/2020.lrec-1.497} {{U}niversal
  {D}ependencies v2: An evergrowing multilingual treebank collection}.
\newblock In \emph{Proceedings of the Twelfth Language Resources and Evaluation
  Conference}, pages 4034--4043, Marseille, France. European Language Resources
  Association.

\bibitem[{Norouzi et~al.(2016)Norouzi, Bengio, Jaitly, Schuster, Wu, Schuurmans
  et~al.}]{norouzi2016reward}
Mohammad Norouzi, Samy Bengio, Navdeep Jaitly, Mike Schuster, Yonghui Wu, Dale
  Schuurmans, et~al. 2016.
\newblock Reward augmented maximum likelihood for neural structured prediction.
\newblock \emph{Advances In Neural Information Processing Systems}, 29.

\bibitem[{Orosz et~al.(2022)Orosz, Sz{\' a}nt{\' o}, Berkecz, Szab{\' o}, and
  Farkas}]{HuSpaCy:2021}
Gy{\"o}rgy Orosz, Zsolt Sz{\' a}nt{\' o}, P{\' e}ter Berkecz, Gerg{\H o}
  Szab{\' o}, and Rich{\' a}rd Farkas. 2022.
\newblock {HuSpaCy: an industrial-strength Hungarian natural language
  processing toolkit}.

\bibitem[{Qi et~al.(2020)Qi, Zhang, Zhang, Bolton, and Manning}]{qi2020stanza}
Peng Qi, Yuhao Zhang, Yuhui Zhang, Jason Bolton, and Christopher~D. Manning.
  2020.
\newblock \href {https://nlp.stanford.edu/pubs/qi2020stanza.pdf} {Stanza: A
  {Python} natural language processing toolkit for many human languages}.
\newblock In \emph{Proceedings of the 58th Annual Meeting of the Association
  for Computational Linguistics: System Demonstrations}.

\bibitem[{{\c{S}}ahin and Steedman(2018)}]{sahin-steedman-2018-data}
G{\"o}zde~G{\"u}l {\c{S}}ahin and Mark Steedman. 2018.
\newblock \href {https://doi.org/10.18653/v1/D18-1545} {Data augmentation via
  dependency tree morphing for low-resource languages}.
\newblock In \emph{Proceedings of the 2018 Conference on Empirical Methods in
  Natural Language Processing}, pages 5004--5009, Brussels, Belgium.
  Association for Computational Linguistics.

\bibitem[{S{\'a}nchez-Cartagena et~al.(2021)S{\'a}nchez-Cartagena,
  Espl{\`a}-Gomis, P{\'e}rez-Ortiz, and
  S{\'a}nchez-Mart{\'\i}nez}]{sanchez-cartagena-etal-2021-rethinking}
V{\'\i}ctor~M. S{\'a}nchez-Cartagena, Miquel Espl{\`a}-Gomis, Juan~Antonio
  P{\'e}rez-Ortiz, and Felipe S{\'a}nchez-Mart{\'\i}nez. 2021.
\newblock \href {https://doi.org/10.18653/v1/2021.emnlp-main.669} {Rethinking
  data augmentation for low-resource neural machine translation: A multi-task
  learning approach}.
\newblock In \emph{Proceedings of the 2021 Conference on Empirical Methods in
  Natural Language Processing}, pages 8502--8516, Online and Punta Cana,
  Dominican Republic. Association for Computational Linguistics.

\bibitem[{Sanfeliu and Fu(1983)}]{6313167}
Alberto Sanfeliu and King-Sun Fu. 1983.
\newblock \href {https://doi.org/10.1109/TSMC.1983.6313167} {A distance measure
  between attributed relational graphs for pattern recognition}.
\newblock \emph{IEEE Transactions on Systems, Man, and Cybernetics},
  SMC-13(3):353--362.

\bibitem[{Sennrich et~al.(2016)Sennrich, Haddow, and
  Birch}]{sennrich-etal-2016-improving}
Rico Sennrich, Barry Haddow, and Alexandra Birch. 2016.
\newblock \href {https://doi.org/10.18653/v1/P16-1009} {Improving neural
  machine translation models with monolingual data}.
\newblock In \emph{Proceedings of the 54th Annual Meeting of the Association
  for Computational Linguistics (Volume 1: Long Papers)}, pages 86--96, Berlin,
  Germany. Association for Computational Linguistics.

\bibitem[{Shi et~al.(2020)Shi, Livescu, and Gimpel}]{shi-etal-2020-role}
Haoyue Shi, Karen Livescu, and Kevin Gimpel. 2020.
\newblock \href {https://doi.org/10.18653/v1/2020.emnlp-main.614} {On the role
  of supervision in unsupervised constituency parsing}.
\newblock In \emph{Proceedings of the 2020 Conference on Empirical Methods in
  Natural Language Processing (EMNLP)}, pages 7611--7621, Online. Association
  for Computational Linguistics.

\bibitem[{Shi et~al.(2021)Shi, Livescu, and
  Gimpel}]{shi-etal-2021-substructure}
Haoyue Shi, Karen Livescu, and Kevin Gimpel. 2021.
\newblock \href {https://doi.org/10.18653/v1/2021.findings-acl.307}
  {Substructure substitution: Structured data augmentation for {NLP}}.
\newblock In \emph{Findings of the Association for Computational Linguistics:
  ACL-IJCNLP 2021}, pages 3494--3508, Online. Association for Computational
  Linguistics.

\bibitem[{Vania et~al.(2019)Vania, Kementchedjhieva, S{\o}gaard, and
  Lopez}]{vania-etal-2019-systematic}
Clara Vania, Yova Kementchedjhieva, Anders S{\o}gaard, and Adam Lopez. 2019.
\newblock \href {https://doi.org/10.18653/v1/D19-1102} {A systematic comparison
  of methods for low-resource dependency parsing on genuinely low-resource
  languages}.
\newblock In \emph{Proceedings of the 2019 Conference on Empirical Methods in
  Natural Language Processing and the 9th International Joint Conference on
  Natural Language Processing (EMNLP-IJCNLP)}, pages 1105--1116, Hong Kong,
  China. Association for Computational Linguistics.

\bibitem[{Varga et~al.(2007)Varga, Hal{\'a}csy, Kornai, Nagy, N{\'e}meth, and
  Tr{\'o}n}]{varga2007parallel}
D{\'a}niel Varga, P{\'e}ter Hal{\'a}csy, Andr{\'a}s Kornai, Viktor Nagy,
  L{\'a}szl{\'o} N{\'e}meth, and Viktor Tr{\'o}n. 2007.
\newblock Parallel corpora for medium density languages.
\newblock \emph{Amsterdam Studies In The Theory And History Of Linguistic
  Science Series 4}, 292:247.

\bibitem[{Vaswani et~al.(2017)Vaswani, Shazeer, Parmar, Uszkoreit, Jones,
  Gomez, Kaiser, and Polosukhin}]{vaswani2017attention}
Ashish Vaswani, Noam Shazeer, Niki Parmar, Jakob Uszkoreit, Llion Jones,
  Aidan~N Gomez, {\L}ukasz Kaiser, and Illia Polosukhin. 2017.
\newblock Attention is all you need.
\newblock In \emph{Advances in neural information processing systems}, pages
  5998--6008.

\bibitem[{Wang and Sennrich(2020)}]{wang-sennrich-2020-exposure}
Chaojun Wang and Rico Sennrich. 2020.
\newblock \href {https://doi.org/10.18653/v1/2020.acl-main.326} {On exposure
  bias, hallucination and domain shift in neural machine translation}.
\newblock In \emph{Proceedings of the 58th Annual Meeting of the Association
  for Computational Linguistics}, pages 3544--3552, Online. Association for
  Computational Linguistics.

\bibitem[{Wang et~al.(2018)Wang, Pham, Dai, and
  Neubig}]{wang-etal-2018-switchout}
Xinyi Wang, Hieu Pham, Zihang Dai, and Graham Neubig. 2018.
\newblock \href {https://doi.org/10.18653/v1/D18-1100} {{S}witch{O}ut: an
  efficient data augmentation algorithm for neural machine translation}.
\newblock In \emph{Proceedings of the 2018 Conference on Empirical Methods in
  Natural Language Processing}, pages 856--861, Brussels, Belgium. Association
  for Computational Linguistics.

\bibitem[{Wei et~al.(2022)Wei, Yu, Hu, Weng, Luo, and
  Jin}]{wei-etal-2022-learning}
Xiangpeng Wei, Heng Yu, Yue Hu, Rongxiang Weng, Weihua Luo, and Rong Jin. 2022.
\newblock Learning to generalize to more: Continuous semantic augmentation for
  neural machine translation.
\newblock In \emph{Proceedings of the 60th Annual Meeting of the Association
  for Computational Linguistics, ACL 2022}.

\bibitem[{Xu et~al.(2016)Xu, Jia, Mou, Li, Chen, Lu, and
  Jin}]{xu-etal-2016-improved}
Yan Xu, Ran Jia, Lili Mou, Ge~Li, Yunchuan Chen, Yangyang Lu, and Zhi Jin.
  2016.
\newblock \href {https://aclanthology.org/C16-1138} {Improved relation
  classification by deep recurrent neural networks with data augmentation}.
\newblock In \emph{Proceedings of {COLING} 2016, the 26th International
  Conference on Computational Linguistics: Technical Papers}, pages 1461--1470,
  Osaka, Japan. The COLING 2016 Organizing Committee.

\bibitem[{Yao et~al.(2022)Yao, Wang, Li, Zhang, Liang, Zou, and
  Finn}]{yao2022improving}
Huaxiu Yao, Yu~Wang, Sai Li, Linjun Zhang, Weixin Liang, James Zou, and Chelsea
  Finn. 2022.
\newblock Improving out-of-distribution robustness via selective augmentation.
\newblock In \emph{International Conference on Machine Learning}, pages
  25407--25437. PMLR.

\bibitem[{Zhao et~al.(2018)Zhao, Wang, Yatskar, Ordonez, and
  Chang}]{zhao-etal-2018-gender}
Jieyu Zhao, Tianlu Wang, Mark Yatskar, Vicente Ordonez, and Kai-Wei Chang.
  2018.
\newblock \href {https://doi.org/10.18653/v1/N18-2003} {Gender bias in
  coreference resolution: Evaluation and debiasing methods}.
\newblock In \emph{Proceedings of the 2018 Conference of the North {A}merican
  Chapter of the Association for Computational Linguistics: Human Language
  Technologies, Volume 2 (Short Papers)}, pages 15--20, New Orleans, Louisiana.
  Association for Computational Linguistics.

\end{thebibliography}

%\appendix

\end{document}